\documentclass[a4paper,twocolumn]{article}

\usepackage[a4paper,margin=2cm,bottom=3cm,columnsep=1cm]{geometry}
\usepackage{graphicx}
\usepackage[hidelinks,bookmarks=false]{hyperref}
\usepackage{multirow}
\usepackage{enumitem}
\usepackage{tabularray}
\usepackage{authblk}
\usepackage[square,numbers]{natbib}
\usepackage{amsmath}

\newcommand\blfootnote[1]{%
  \begingroup
  \renewcommand\thefootnote{}\footnote{#1}%
  \addtocounter{footnote}{-1}%
  \endgroup
}

\title{Multimodal Large Language Models\\for Real-Time Situated Reasoning}

\author{Giulio Antonio Abbo}
\author{Senne Lenaerts}
\author{Tony Belpaeme}
\affil{\textit{IDLab-AIRO}, \textit{Ghent University -- imec}, Ghent, Belgium}
\affil{\texttt{giulioantonio.abbo@ugent.be}}

\date{}

\begin{document}

\maketitle

\begin{abstract}
In this work, we explore how multimodal large language models can support real-time context- and value-aware decision-making. To do so, we combine the GPT-4o language model with a TurtleBot~4 platform simulating a smart vacuum cleaning robot in a home. The model evaluates the environment through vision input and determines whether it is appropriate to initiate cleaning. The system highlights the ability of these models to reason about domestic activities, social norms, and user preferences and take nuanced decisions aligned with the values of the people involved, such as cleanliness, comfort, and safety. We demonstrate the system in a realistic home environment, showing its ability to infer context and values from limited visual input. Our results highlight the promise of multimodal large language models in enhancing robotic autonomy and situational awareness, while also underscoring challenges related to consistency, bias, and real-time performance.
\end{abstract}

\blfootnote{Submitted to the \textit{interactivity} track of the 21st ACM/IEEE International Conference on Human-Robot Interaction on December 2025, accepted January 2026.}

%%%%%%%%%%%%%%%%%%%%%%%%%%%%%%%%%%%%%%%%%%%%%%%%%%%%%%%%%%%%%%%%%%%%%%%%%%%%%%%%
\section{Introduction}

Strong human-robot interaction relies on effective environmental modelling, as this provides the foundation for informed decision-making.
Robots must dynamically perceive and interpret their surroundings, including objects, obstacles, and human actions, while accounting for uncertainties such as sensor limitations and unexpected changes.
Without a robust and adaptive model, robots risk inefficiency or even unsafe behaviour, undermining their utility in real-world applications.

\begin{figure}[t!]
  \centering
  \includegraphics[width=0.5\columnwidth,alt={The TurtleBot robot, which looks like a vacuum cleaner robot, on the floor, and a dog next to it.}]{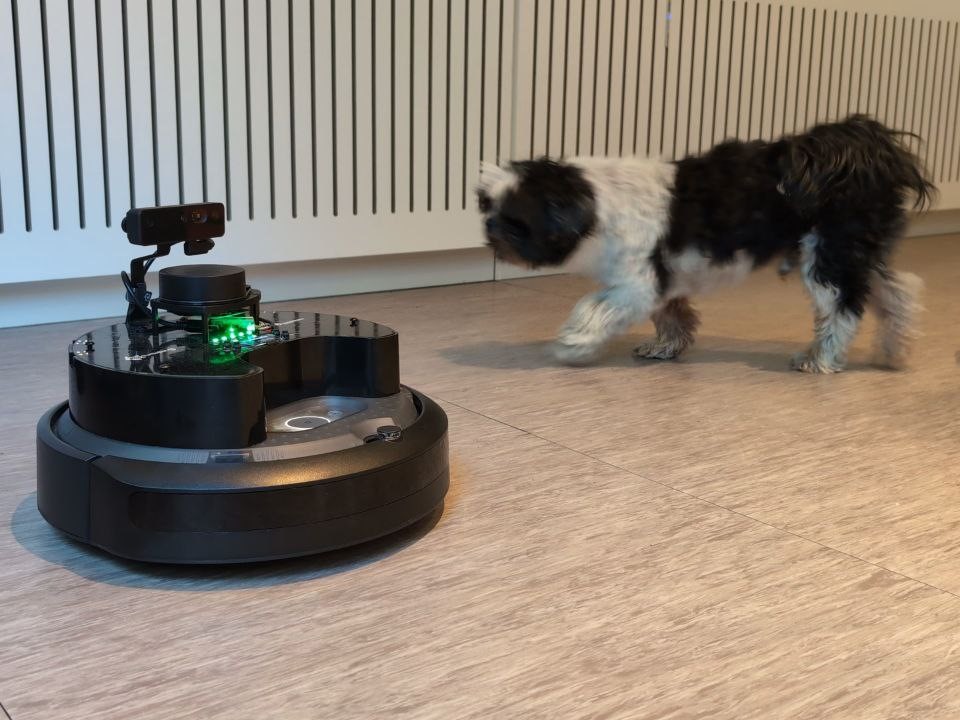}%
  \includegraphics[width=0.5\columnwidth,alt={A low-angle view of a living room, with the bottom part of a couch and the legs of a person sitting on it.}]{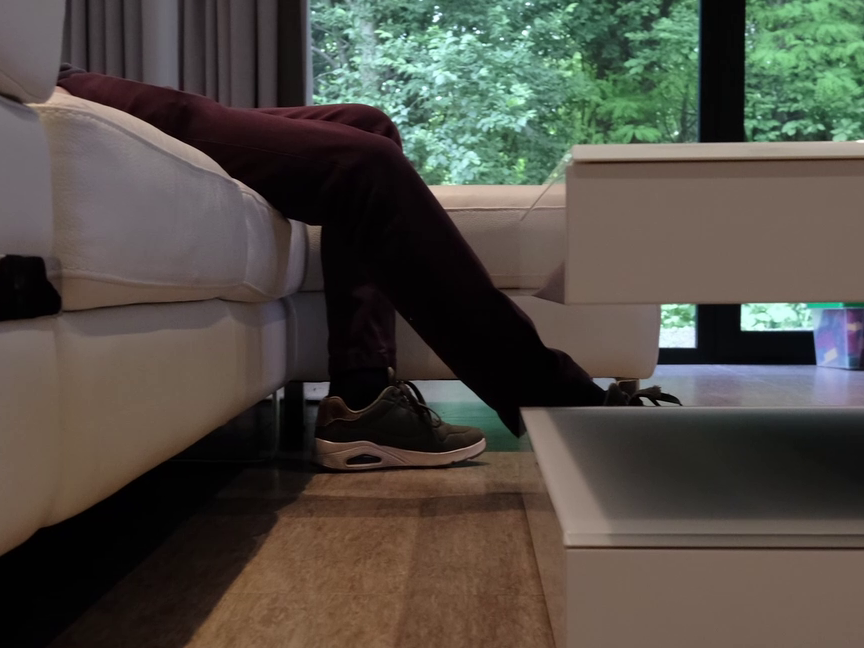}
  \caption{The robot deployed in a living room, and an image from the robot's perspective, showing the unusual angle and the partially obstructed view.}
  \label{fig:observations}
\end{figure}

Yet, modelling real-life environments presents significant challenges.
For instance, in a home, the complexity of unstructured scenarios, with variables such as values and personal preferences, family dynamics, and traditions, can quickly overwhelm traditional approaches.
Scaling these models to handle diverse, dynamic settings often demands meticulously handcrafted rules, which reduce the flexibility and scalability of these approaches.

Recent advancements in Large Language Models offer a compelling opportunity to equip household robots with a higher level of decision-making.
These models have demonstrated impressive capabilities not only in mimicking~\cite{illusion-thinking} contextual reasoning~\cite{plaat2024reasoninglargelanguagemodels} and common sense understanding, but also in guiding robots' behaviour~\cite{geminirobotics}.
When combined with multimodal inputs, such as images or sensor data, multimodal LLMs can interpret a scene and evaluate whether specific actions align with human preferences, norms, and routines~\cite{vision-enabled-dialogue,visionllms}.
Furthermore, the text-based nature of these approaches produces a human-readable decision trace that explains the rationale behind each decision taken, and it can easily be integrated with additional users' indications and corrections.
However, despite their potential, multimodal LLMs as decision-makers might struggle with real-time performance.
Indeed, it is necessary to balance the quality and completeness of the reasoning process result with its duration.
Without careful prompting, LLMs can be inconsistent in their decision outcomes, which is undesirable, especially in social robot applications.

This work explores how multimodal LLMs can be used to reason about values and preferences in real time.
To do so, we combine an LLM-based reasoning framework with a TurtleBot 4 robot acting as a value-aware vacuum cleaner in a home setting.

Autonomous vacuum cleaners have become a common presence in modern households, offering convenience through scheduled or reactive cleaning routines.
These robots typically rely on a combination of sensors, pre-programmed logic, and, in some cases, basic learning algorithms to navigate environments, avoid obstacles, and perform their cleaning tasks.
While increasingly effective in manoeuvring through rooms, these systems still fall short in understanding the context in which cleaning occurs.
In real-world scenarios, determining whether or not cleaning is appropriate involves more than obstacle avoidance and scheduling.
For instance, a person watching a movie in the living room might prefer silence, while a recently used kitchen might benefit from immediate cleaning to avoid insects taking over.
These nuanced decisions depend not only on detecting objects or motion, but also on understanding activities, routines, and values that govern comfort in a shared space.
A robot that can \textit{reason} about ongoing activities and routines would be better aligned with users' expectations and preferences.

In our implementation, the robot uses image input to interpret its environment and make contextually appropriate decisions about whether to clean, wait, or return to its docking station.
We observe the system in action in a real home environment, identifying strengths and limitations in the robot's contextual understanding and value alignment.

Through this work, we aim to demonstrate that multimodal LLMs provide a promising foundation for more intelligent, empathetic, and situationally aware household robots.

%%%%%%%%%%%%%%%%%%%%%%%%%%%%%%%%%%%%%%%%%%%%%%%%%%%%%%%%%%%%%%%%%%%%%%%%%%%%%%%%
\section{Implementation}

The robot used is a TurtleBot 4, an open-source platform for research created by Open Robotics and Clearpath Robotics\footnote{\url{https://clearpathrobotics.com/turtlebot-4/}}.
The robot is built on top of the iRobot \textit{CREATE 3} mobile base, which is based on the chassis of a robot vacuum cleaner.
The robot has a rich sensor suite, with our research relying on the \textit{OAK-D-PRO} RGBD camera to collect images from the robot's point of view.

The system driving the robot has three states: \textit{observation}, \textit{cleaning}, and \textit{docking} mode, with the VLM (Vision Language Model) deciding when to transition from one state to another.
 
In \textbf{observation mode}, the robot surveys its environment to collect a series of images. From these salient features are extracted and passed to the model, which decides which action to take.
Specifically, the robot starts rotating on itself and collects a series of 10 images at regular intervals, giving a 180-degree overview of the surrounding space at one frame per second.
A VLM is used to extract key elements from the image, focusing on the presence of people or pets, human activities, and contextual clues.
The list of relevant elements and the original images are then used by the same VLM to perform the decision-making process.
The system prompt that instructs the model how to behave is composed of three parts.
First, it defines a role: ``a value-aware vacuum cleaner''; then it defines the objective of ``maintain cleanliness while respecting the homeowner’s values, ensuring a comfortable and harmonious living environment.''
Finally, it describes the three modes in detail, including the capabilities and behaviour of the robot in each mode.
The model is prompted to reason step-by-step on the following aspects: value alignment, ``creating a safe and enjoyable environment for the house owners'', time context -- with the current time provided in the prompt -- the action choice and its consequences, the rationale, and how the decision reflects the robot's core purpose of cleaning.
The reasoning trace is then fed back into the model to take a final decision, and, finally, the entire process is summarised into a short, user-friendly paragraph to add a level of explainability.
We relied on OpenAI's GPT-4o model, which accepts both image and text inputs and produces text-based outputs.

In \textbf{cleaning mode}, the robot simulates a vacuum cleaning robot by driving in a straight line until the proximity sensors of the robot detect a nearby object, upon which the robot slows down. If a collision is detected, it turns in a random direction and resumes cleaning.
In the meantime, the robot collects 3 images every 0.5 seconds and uses the VLM to interpret the environment and evaluate whether any values are potentially violated.
This determines whether the current cleaning task should be interrupted and the robot should transition into observation mode using a reasoning process similar to the previous mode.

\textbf{Docking mode} instructs the robot to return to its charging station.
While in the current implementation this mode stops the robot, in the future it could be used to instruct the robot to go to another room, if the current room is busy, before finally going to the charging station.

ROS 2, the robot operating system~\cite{ros2}, is used to implement the functionalities described.
This middleware handles the communication between specialised nodes which can run on different machines.
The TurtleBot 4 comes with several nodes already running, to collect images from the camera and controlling the robot.
The system presented runs as a Python node on another computer, connected to the same wireless network as the robot.
In addition, a graphical interface developed with PyQT5 displays the reasoning process and allows control of the robot.

%%%%%%%%%%%%%%%%%%%%%%%%%%%%%%%%%%%%%%%%%%%%%%%%%%%%%%%%%%%%%%%%%%%%%%%%%%%%%%%%
\section{Observations}

Evaluating the value alignment of the system responses depends strongly on the subjective preferences of the people involved.
As an initial evaluation, we opted for a qualitative observation of the system responses -- without a quantitative evaluation -- first using a small dataset, after which the system was evaluated in a realistic environment.

To observe the VLM's behaviour without the additional complications of the robot's hardware, the system was used with a set of images instead of the live feed form the robot's camera.
These images are manually chosen from the YouHome Activities of Daily Living dataset~\cite{youhome}, which comprises 31 common daily activities from 20 users, in different areas of a home.
We chose images that could allow observing how well the model detects people and infers which activities they are partaking in.
In addition, we also tried scenarios in which the room appears empty, or someone enters and then leaves the room in a quick succession.

By deploying the system in a home, we obtained a more comprehensive idea of the overall system performance.
We put the robot in a living room, and observed its decision-making process under three conditions: when someone is watching a movie, using a phone, and around a pet dog.

In the first case, shown in Figure~\ref{fig:observations}, the robot was positioned near a sofa, on top of which a person was sitting watching a television show.
The robot, even from the low and partially obstructed view angle, was able to correctly recognise the activity, and decide to defer the cleaning using the noises produced as a justification.
We then used the same setting, but with the person focused on their phone and no movie playing.
In this case, the VLM reasoned that using the phone does not require silence, and to balance the users' comfort with the purpose of cleaning, it decided to start cleaning.
Finally, when a dog was introduced to the robot, the system noted that the pet appeared to be moving a lot in its  vicinities, and thus cleaning was not advisable for the pet's physical safety.
The system also reflected that the loud noises could scare the pet, showing that its reasoning capabilities can take into account the mental wellbeing of those involved as well as their safety.

Our work has several limitations.
We tested and decided not to adopt reasoning models and several prompting techniques, including learning from examples, due to the additional delays that longer prompts imply.
In the future, we intend to further explore these possibilities, including the use of locally hosted models, which would alleviate the privacy issues tied to using pictures from within the home environment.
The inherent bias~\cite{concerns-and-values} of the multimodal LLMs needs to be further studied and mitigated using additional information in the prompt, such as users' habits and preferences, which can be expressed by the users themselves through voice or text and integrated into the prompt.

%%%%%%%%%%%%%%%%%%%%%%%%%%%%%%%%%%%%%%%%%%%%%%%%%%%%%%%%%%%%%%%%%%%%%%%%%%%%%%%%
\section{Conclusion}

We tested how well state-of-the-art multimodal language models can reason on personal preferences and values in real time. To do so, we presented an implementation of a vacuum cleaning robot based on the TurtleBot 4 platform and GPT-4o.
The robot collects images from its camera and uses them to decide whether to start cleaning or not.
We deployed the robot in a living room and observed its behaviour.
Overall, the language model, despite the unusual point of view, the partially obstructed images, and the limited time resolution, was successful in inferring the activities taking place in the room, in using this information to reason on the values involved, and balancing the values of cleanliness and respect for the people involved.

This work aims to demonstrate that large language models, when integrated into embodied systems, go beyond task execution and begin to reason about the values and context in which their actions take place, marking a significant step toward more socially aware and adaptable domestic robots.

\section*{Acknowledgements}
Funded by the Horizon Europe VALAWAI project (grant agreement number 101070930).

\bibliographystyle{ACM-Reference-Format}
\bibliography{bibliography}

@inproceedings{youhome,
author={Pan, Junhao and Yuan, Zehua and Zhang, Xiaofan and Chen, Deming},
booktitle={2022 IEEE International Symposium on Smart Electronic Systems (iSES)}, 
title={YouHome System and Dataset: Making Your Home Know You Better}, 
year={2022},
volume={},
number={},
pages={414-420},
doi={10.1109/iSES54909.2022.00091}
}

@misc{geminirobotics,
title={Gemini Robotics: Bringing AI into the Physical World}, 
author={Gemini Robotics Team and Saminda Abeyruwan and Joshua Ainslie and Jean-Baptiste Alayrac and Montserrat Gonzalez Arenas and Travis Armstrong and Ashwin Balakrishna and Robert Baruch and Maria Bauza and Michiel Blokzijl and Steven Bohez and Konstantinos Bousmalis and Anthony Brohan and Thomas Buschmann and Arunkumar Byravan and Serkan Cabi and Ken Caluwaerts and Federico Casarini and Oscar Chang and Jose Enrique Chen and Xi Chen and Hao-Tien Lewis Chiang and Krzysztof Choromanski and David D'Ambrosio and Sudeep Dasari and Todor Davchev and Coline Devin and Norman Di Palo and Tianli Ding and Adil Dostmohamed and Danny Driess and Yilun Du and Debidatta Dwibedi and Michael Elabd and Claudio Fantacci and Cody Fong and Erik Frey and Chuyuan Fu and Marissa Giustina and Keerthana Gopalakrishnan and Laura Graesser and Leonard Hasenclever and Nicolas Heess and Brandon Hernaez and Alexander Herzog and R. Alex Hofer and Jan Humplik and Atil Iscen and Mithun George Jacob and Deepali Jain and Ryan Julian and Dmitry Kalashnikov and M. Emre Karagozler and Stefani Karp and Chase Kew and Jerad Kirkland and Sean Kirmani and Yuheng Kuang and Thomas Lampe and Antoine Laurens and Isabel Leal and Alex X. Lee and Tsang-Wei Edward Lee and Jacky Liang and Yixin Lin and Sharath Maddineni and Anirudha Majumdar and Assaf Hurwitz Michaely and Robert Moreno and Michael Neunert and Francesco Nori and Carolina Parada and Emilio Parisotto and Peter Pastor and Acorn Pooley and Kanishka Rao and Krista Reymann and Dorsa Sadigh and Stefano Saliceti and Pannag Sanketi and Pierre Sermanet and Dhruv Shah and Mohit Sharma and Kathryn Shea and Charles Shu and Vikas Sindhwani and Sumeet Singh and Radu Soricut and Jost Tobias Springenberg and Rachel Sterneck and Razvan Surdulescu and Jie Tan and Jonathan Tompson and Vincent Vanhoucke and Jake Varley and Grace Vesom and Giulia Vezzani and Oriol Vinyals and Ayzaan Wahid and Stefan Welker and Paul Wohlhart and Fei Xia and Ted Xiao and Annie Xie and Jinyu Xie and Peng Xu and Sichun Xu and Ying Xu and Zhuo Xu and Yuxiang Yang and Rui Yao and Sergey Yaroshenko and Wenhao Yu and Wentao Yuan and Jingwei Zhang and Tingnan Zhang and Allan Zhou and Yuxiang Zhou},
year={2025},
eprint={2503.20020},
archivePrefix={arXiv},
url={https://arxiv.org/abs/2503.20020}, 
}

@misc{illusion-thinking,
title={The Illusion of Thinking: Understanding the Strengths and Limitations of Reasoning Models via the Lens of Problem Complexity}, 
author={Parshin Shojaee and Iman Mirzadeh and Keivan Alizadeh and Maxwell Horton and Samy Bengio and Mehrdad Farajtabar},
year={2025},
eprint={2506.06941},
archivePrefix={arXiv},
primaryClass={cs.AI},
url={https://arxiv.org/abs/2506.06941}, 
}

@article{ros2,
author = {Steven Macenski  and Tully Foote  and Brian Gerkey  and Chris Lalancette  and William Woodall },
title = {Robot Operating System 2: Design, architecture, and uses in the wild},
journal = {Science Robotics},
volume = {7},
number = {66},
pages = {eabm6074},
year = {2022},
doi = {10.1126/scirobotics.abm6074},
}

@InProceedings{visionllms,
author="Abbo, Giulio Antonio
and Belpaeme, Tony",
editor="Osman, Nardine
and Steels, Luc",
title="Vision Language Models as Values Detectors",
booktitle="Value Engineering in Artificial Intelligence",
year="2025",
publisher="Springer Nature Switzerland",
address="Cham",
pages="76--86",
isbn="978-3-031-85463-7"
}

@misc{plaat2024reasoninglargelanguagemodels,
title={Reasoning with Large Language Models, a Survey}, 
author={Aske Plaat and Annie Wong and Suzan Verberne and Joost Broekens and Niki van Stein and Thomas Back},
year={2024},
eprint={2407.11511},
archivePrefix={arXiv},
primaryClass={cs.AI},
url={https://arxiv.org/abs/2407.11511}, 
}

@inproceedings{vision-enabled-dialogue,
author = {Abbo, Giulio Antonio and Belpaeme, Tony},
title = {I Was Blind but Now I See: Implementing Vision-Enabled Dialogue in Social Robots},
year = {2025},
publisher = {IEEE Press},
booktitle = {Proceedings of the 2025 ACM/IEEE International Conference on Human-Robot Interaction},
pages = {1176–1180},
numpages = {5},
location = {Melbourne, Australia},
series = {HRI '25}
}

@Article{concerns-and-values,
author={Abbo, Giulio Antonio
and Belpaeme, Tony
and Spitale, Micol},
title={Concerns and Values in Human-Robot Interactions: A Focus on Social Robotics},
journal={International Journal of Social Robotics},
year={2026},
month={Jan},
day={14},
volume={18},
number={1},
pages={4},
issn={1875-4805},
doi={10.1007/s12369-025-01351-1},
url={https://doi.org/10.1007/s12369-025-01351-1}
}

\end{document}